\pdfoutput=1

\documentclass[11pt]{article}

\usepackage[]{ACL2023}

\usepackage{balance} 
\usepackage{amsmath}
\usepackage{graphicx}
\usepackage{multirow}
\usepackage{colortbl}
\usepackage{booktabs}

\usepackage{times}
\usepackage{latexsym}

\usepackage[T1]{fontenc}

\usepackage[utf8]{inputenc}

\usepackage{microtype}

\usepackage{inconsolata}

\definecolor{lightskyblue}{rgb}{0.53, 0.81, 0.98}
\definecolor{lightgray}{rgb}{0.53, 0.81, 0.98}

\title{The Factuality of Large Language Models in the Legal Domain}

\author{
Rajaa EL HAMDANI{\normalfont ,}\textsuperscript{1}
Thomas Bonald{\normalfont ,}\textsuperscript{1}
Fragkiskos Malliaros{\normalfont ,}\textsuperscript{2,3}
Nils Holzenberger{\normalfont ,}\textsuperscript{1}
Fabian Suchanek\textsuperscript{1}
\\
\textsuperscript{1}Télécom Paris, Institut Polytechnique de Paris \\
\textsuperscript{2}Université Paris-Saclay, CentraleSupélec,
\textsuperscript{3}INRIA \\
\href{mailto:rajaa.elhamdani@telecom-paris.fr}{\texttt{{\color{black} rajaa.elhamdani@telecom-paris.fr}}},
\href{mailto:thomas.bonald@telecom-paris.fr}{\texttt{{\color{black} thomas.bonald@telecom-paris.fr}}} \\
\href{mailto:fragkiskos.malliaros@centralesupelec.fr}{\texttt{{\color{black} fragkiskos.malliaros@centralesupelec.fr}}}\\
\href{mailto:nils.holzenberger@telecom-paris.fr}{\texttt{{\color{black} nils.holzenberger@telecom-paris.fr}}},
\href{mailto:fabian.suchanek@telecom-paris.fr}{\texttt{{\color{black} fabian.suchanek@telecom-paris.fr}}}
}

\begin{document}
\maketitle
\begin{abstract}
  This paper investigates the factuality of large language models (LLMs) as knowledge bases in the legal domain, in a realistic usage scenario: we allow for acceptable variations in the answer, and let the model abstain from answering when uncertain.
  First, we design a dataset of diverse factual questions about case law and legislation.
  We then use the dataset to evaluate several LLMs under different evaluation methods, including exact, alias, and fuzzy matching.
  Our results show that the performance improves significantly under the alias and fuzzy matching methods.
  Further, we explore the impact of abstaining and in-context examples, finding that both strategies enhance precision.
  Finally, we demonstrate that additional pre-training on legal documents, as seen with SaulLM, further improves factual precision from 63\% to 81\%.
\end{abstract}

\section{Introduction}

Language models (LMs) store knowledge from pre-training documents, allowing them to be queried with natural language (NL) rather than a formal language, unlike structured knowledge bases (KBs) \cite{petroni_language_2019, chang_language_2024}. This natural language interface democratizes access to knowledge, especially in domains where users may not master querying structured KBs \cite{hendrix_developing_1978}. In the legal domain, this includes law experts (scholars, judges, lawyers, paralegals) and laymen seeking legal information \cite{boniol2020performance}. LMs can acquire and store more facts than KBs, as most knowledge is in NL documents. This reduces the cost of manually populating KBs, as LMs automatically store knowledge from pre-training documents without human supervision \cite{alkhamissi_review_2022}. While Wikidata is the largest structured KB, it lacks comprehensive domain-specific knowledge \cite{weikum_machine_2020}. In contrast, Wikipedia articles offer this knowledge, which LMs can store through pre-training.
\begin{figure}[t]
    \centering
    \fontsize{6pt}{8pt}\selectfont
    \noindent\fcolorbox{white}{lightskyblue!40}{
    \begin{minipage}{0.95\columnwidth}
Your task is to answer legal questions. Provide a concise answer without an explanation.\\ 
If you don't know the answer, respond with "I don't know". Examples:\\
\textbf{Question:} Who wrote the majority opinion of the case "Trump v. Hawaii"?\\
\textbf{Answer:} John Roberts\\
\textbf{Question:} Who wrote the majority opinion of the case "McGirt v. Oklahoma"?\\
\textbf{Answer:} Neil Gorsuch\\
\textbf{Question:} Who wrote the majority opinion of the case "Limtiaco v. Camacho"?\\
\textbf{Answer:} Clarence Thomas\\
\textbf{Question:} Who wrote the majority opinion of the case "R. v. Morgentaler"?\\
\textbf{Answer:} Brian Dickson\\
\textbf{Question:} Who wrote the majority opinion of the case "CREW v. Trump"?\\
\textbf{Answer:} Pierre N. Leval\\
\textbf{Question:} Who wrote the majority opinion of the case "Lau v. Nichols"?
    \end{minipage}
    }
    \caption{Example of prompt for the relation \textit{majority opinion by}. The few shot examples are tailored for the relation \textit{majority opinion by} and the subject's class \textit{legal case}.}
    \label{fig:prompt_case}
\end{figure}


The concept of language models as knowledge bases (LM-as-KB) was introduced by \cite{petroni_language_2019}, demonstrating on the LAMA dataset that BERT retrieves facts better than traditional knowledge bases. Later research \cite{heinzerling_language_2021} showed that large language models (LLMs) with more than 10B parameters can store all of Wikidata and retrieve facts using NL queries, though the accuracy can be sensitive to minor query phrasing changes. Additionally, fine-tuning LMs or using query variants improves their robustness \cite{jiang_how_2020, heinzerling_language_2021, adolphs_how_2021}.
LAMA \cite{petroni_language_2019} was the first dataset to benchmark LM-as-KB, limited to one-token facts for simplicity. Subsequent benchmarks include more complex queries encompassing various types of knowledge \cite{bisk_piqa_2019, peng_copen_2022, lin_birds_2020, misra_comps_2023, elazar_measuring_2021}. Among these, KAMEL \cite{kalo_kamel_2022} proved to be harder than LAMA because the dataset contains multi-token facts, facts with literal values, a broader range of knowledge, and relations of higher cardinality.

Early research on LM-as-KB primarily evaluated smaller models like BERT. However, the advent of LLMs like OpenAI's GPT, Google's PALM, and Meta's Llama \cite{openai_gpt-4_2024, chowdhery_palm_2022, touvron_llama_2023}, revolutionized the field. These models exhibit emergent abilities useful for LM-as-KB, including few-shot prompting \cite{brown_language_2020} and calibration \cite{wei2022emergent}. LLMs can learn new tasks through few-shot prompting without fine-tuning. Since LM-as-KB performance improves with prompt fine-tuning \cite{qin_learning_2021}, including query examples in the prompt can enhance the factuality of LMs. Additionally, well-calibrated models can predict the correctness of their own responses, enabling self-evaluation of answers to factual queries \cite{kadavath_language_2022, geng_survey_2024}.

Despite their capabilities, LLMs are prone to \emph{hallucination} \cite{ji_survey_2023, suchanek2023knowledge}, producing factually incorrect answers in a disturbingly self-confident tone. While many studies have examined the factual accuracy of LLMs \cite{openai_gpt-4_2024, wei_long-form_2024, muhlgay_generating_2024, hu_towards_2024, si_prompting_2023, min_factscore_2023, manakul_selfcheckgpt_2023, mallen_when_2023}, few focused on domain-specific knowledge. Domain-specific knowledge poses unique challenges, such as specific named entities that are unique to the domain (e.g., lawyers, judges) and that do not appear in abundance on the Internet (and hence in training data); relations that may not even be known to the general public (e.g. the judge who issues the majority opinion of a case); facts that are specific to the domain (e.g., legal rules); and finally ways of phrasing that are typical for the jargon, but unusual on the Web as a whole (e.g., concurring or dissenting opinions). However, domain-specific applications have huge potential, aiding users in areas beyond one-size-fits-all products.

In this paper, we focus on factual answers in the legal domain, motivated by both the risks and opportunities of using LLMs in this field \cite{Bommasani2021FoundationModels}. While benchmarks like LawBench \cite{fei_lawbench_2023} and LegalBench \cite{guha_legalbench_2023} evaluate LLMs on various legal tasks, LawBench is specific to Chinese law, and LegalBench lacks a legal knowledge task, limiting their use for probing LLMs on legal facts. The legal domain is particularly sensitive to hallucinations, as incorrect information can lead to harmful decisions \cite{dahl_large_2024}. Previous work used strict evaluation criteria, accepting answers only if identical to the ground truth, ignoring valid response variations ("Samuel A. Alito, Jr" is as valid an answer as "Justice Alito"). Additionally, earlier studies forced LLMs to provide answers, increasing hallucinations. In contrast, our approach allows models to abstain when unsure, reducing hallucinations. Accounting for acceptable variations in the phrasing of the answer and allowing the model to abstain from answering brings our evaluation closer to a realistic human use of an \mbox{LLM as a KB.}

In our work, we address four key research questions: 
\begin{enumerate}
\item Are all LLMs equally affected by the shortcomings of exact matching-based evaluation, or are some more penalized? 
\item Can an LLM abstain from generating incorrect answers? 
\item Does few-shot prompting increase the factuality of LLMs? 
\item Does training on legal documents improve the factuality of an LLM?
\end{enumerate}

We present the methodology in Section \ref{sec:method}, the results  in Section \ref{sec:result}, and our conclusions in Section \ref{sec:conclusion}.

\section{Methodology}
\label{sec:method}

\subsection{Dataset}
Legal knowledge is primarily stored in legislation and legal cases~\cite{raz_authority_2002}. This study focuses on atomic information, such as the jurisdiction of a case or legislation. Even if most real-world queries may not be about atomic facts, these atomic facts are the fundamental building blocks of actionable legal information. Therefore, it is crucial to first ensure that the LLM is not hallucinating atomic facts before asking for the relevant legal rule to a legal dispute. We use Wikidata \cite{vrandecic_wikidata_2014} to create a benchmark dataset of atomic legal facts on legislation and legal cases.

Manually navigating Wikidata's taxonomy to identify relevant types is impractical due to its complexity \cite{suchanek2023integrating}. Instead, we leverage the Wikipedia Category Graph (WCG), which organizes articles by categories to facilitate navigation. We identify legal articles within two hierarchical steps of the ``Law'' category in the WCG and retrieve their corresponding Wikidata items. From these items, we construct a legal taxonomy and query Wikidata's SPARQL endpoint for more relevant items. We refine the dataset by manually selecting relations specific to the legal domain and augmenting it with facts from Wikipedia infoboxes \cite{lehmann_dbpedia_2015}. For evaluation, we manually create question templates for each relation to query LLMs via natural language, resulting in 8,920 question-answer pairs covering various relations (see Table \ref{tab:relation_counts}).\footnote{Code and dataset available at \href{https://github.com/Rajjaa/LexFact}{https://github.com/Rajjaa/LexFact}}

\begin{table}[t]
\centering
\begin{tabular}{lr}
\hline
Relation & count \\
\hline
applies to jurisdiction & 3,155 \\
majority opinion by & 1,704 \\
legislated by & 1,345 \\
signatory & 817 \\
language of work or name & 730 \\
amended by & 518 \\
plaintiff & 342 \\
defendant & 309 \\
\hline
Total & 8,920 \\
\hline
\end{tabular}
\caption{Dataset statistics.}
    \label{tab:relation_counts}
\end{table}

\subsection{Models}
Large models are costly to deploy on-premise. We argue that most, if not all, legal tech applications involve sensitive personal data that cannot be transferred to third-party APIs. Thus, we limit experiments to open-source models with fewer than 8B parameters, runnable on a local machine. We select 7 models among the highly ranked in the LMSYS Chatbot Arena Leaderboard \cite{chiang2024chatbot}: \texttt{Gemma-2B}, \texttt{Gemma-7B}, \texttt{Llama-2-7B}, \texttt{Llama-3-8B}, \texttt{Mistral-7B}, \texttt{Phi-3-min-4k}, \texttt{Recurrent\-Gemma-2B} \cite{team2024gemma, botev2024recurrentgemma, touvron_llama_2023}. We  add \texttt{SaulLM} \cite{colombo_saullm-7b_2024}, which is the result of further training  \texttt{Mistral-7B} on legal corpora. We use the instruction-tuned variant of each model. 

\subsection{Prompt strategy}
We experiment with zero-shot and few-shot prompts to query the models. For a question $q$, related to a subject-relation pair $(s,r)$, the corresponding prompt contains 5 in-context examples\footnote{We experimented with one example. For most models, one example is not enough to learn the correct formatting of the answer.} depending on the relation $r$ and the class of the subject, say class $c$. In-context examples are question-answer pairs about subjects of class $c$ and relation $r$. We do not query the model for questions used as in-context examples to avoid data leakage. Figure \ref{fig:prompt_case} shows an example prompt. 
We also instruct the model to answer ``I don't know'' if it cannot answer the question correctly.

\subsection{Evaluation methods}
We are evaluating the LLM as a KB. The correctness and coverage of a KB are usually evaluated by precision $P$ and recall $R$ \cite{weikum_machine_2020}, respectively. For a given KB containing a set of statements $S$ and a set of true statements $T$ from the ground truth, they are defined by:
$$P = \frac{S \bigcap T}{S} \quad \text{and}\quad   R=\frac{S \bigcap T}{T}.$$
In our setting of LLM-as-KB, $T$ is the set of  all question-answer pairs (those of Table \ref{tab:relation_counts}),   $S$ is the set of questions answered by the LLM,\footnote{The LLM might abstain from answering a question.} and $S\cap T$ is the set of correct answers. Thus, the corresponding precision and recall are given by: 
$$ \text{P}_{LLM} = \frac{|\text{correct answers}|}{|\text{answered questions}|}$$
$$ \text{R}_{LLM} = \frac{|\text{correct answers}|}{|\text{all questions}|}$$

Most articles evaluating LLMs consider an answer correct only if it exactly matches the ground truth \cite{dahl_large_2024, petroni_language_2019, chang_survey_2024, hu_towards_2024, yu_kola_2023}. However, LLMs can generate a fact in multiple  forms and tend to answer verbosely, often providing explanations along with the answer. This behavior penalizes LLMs when evaluated with exact matching.

Each Wikidata item has a main surface form called a label and alternative names known as \emph{aliases}. For example, item Q30 has the label ``United States of America'' and aliases like ``U.S.A.'' and ``America''.  We consider the following options for declaring an answer correct:
\begin{enumerate}
\item[(EM)] {\bf Exact matching.} The answer matches the label.
\item[(AM)] {\bf Alias matching.}
  The answer matches the label or any of its aliases.
wI\item[(FM)] {\bf Fuzzy matching.}
 The answer contains the label or any of its aliases.
\end{enumerate}
Fuzzy matching accounts for correct but verbose answers. It is prone to errors, however. Take, for instance, the question ``What is the legislation of the case `Rummel v. Estelle'?'', whose correct answer is ``United States''. The  answer ``The case `Rummel v. Estelle' applies to the state of Louisiana, United States'' is false but contains the label of the true answer. We manually inspect such cases and implement post-processing rules for the answers to enhance the accuracy of fuzzy matching. 





\section{Results}
\label{sec:result}

\subsection{Exact matching underestimates the performance of LLM-as-KB}

The performance of the models, shown in Table \ref{tab:micro_scores}, increases significantly when evaluated with alias matching (AM) and fuzzy matching (FM), compared to exact matching (EM). This indicates that LLMs often generate correct answers in varied surface forms that are not captured by exact matching alone. For instance, \texttt{SaulLM} in few-shot mode and allowed to abstain increased in precision from 36\% (EM) to 81\% (FM).
The ranking of the models changed notably depending on the evaluation method. Under exact matching and alias matching, \texttt{Mistral-7B} performed poorly, ranking last with 8\% precision (EM). However, under fuzzy matching, \texttt{Mistral-7B}'s improved significantly, ranking third with a precision of 63\%.

\begin{table}[t]
\fontsize{11pt}{12.5pt}\selectfont
\resizebox{\columnwidth}{!}{
\begin{tabular}{lll|cc|cc|cc|c}
\toprule 
 \multicolumn{3}{c}{ } & \multicolumn{2}{c}{EM} & \multicolumn{2}{c}{AM} & \multicolumn{2}{c}{FM} &  \\
Model & Prompt & Abstain & P & R & P & R & P & R & Abstain Rate \\
\midrule 

 \rowcolor{lightskyblue!30} \cellcolor{white}\multirow[t]{4}{*}{\texttt{SaulLM}} &\multirow[t]{2}{*}{Zero-shot} & True & \phantom{0}3.7 & \phantom{0}0.5 & \phantom{0}4.4 & \phantom{0}0.6 & \underline{68.3} & \phantom{0}9.9 & 85.6 \\
  \rowcolor{lightskyblue!30} \cellcolor{white}&   & False & \phantom{0}0.8 & \phantom{0}0.8 & \phantom{0}2.2 & \phantom{0}2.2 & 60.9 & 60.9 & \phantom{0}0.1 \\
\cline{2-10} 
 & \multirow[t]{2}{*}{Few-shot} & True & \textbf{35.6 }& 21.4 & \textbf{73.2} & 44.1 & \textbf{81.2} & 48.9 & 39.8 \\
 &  & False & 30.0 & 30.0 & 57.6 & \textbf{57.6} & 62.9 & \textbf{62.9} & \phantom{0}0.0 \\
\midrule
 \rowcolor{lightskyblue!30} \cellcolor{white}\multirow[t]{4}{*}{\texttt{Mistral}} & \multirow[t]{2}{*}{Zero-shot} & True & \phantom{0}0.1 & \phantom{0}0.0 & \phantom{0}0.1 & \phantom{0}0.1 & 65.3 & 55.6 & 14.9 \\
  \rowcolor{lightskyblue!30} \cellcolor{white}&  & False & \phantom{0}0.0 & \phantom{0}0.0 & \phantom{0}0.0 & \phantom{0}0.0 & 58.5 & 57.7 & \phantom{0}1.4 \\
\cline{2-10}
 & \multirow[t]{2}{*}{Few-shot} & True & \phantom{0}8.1 & \phantom{0}7.6 & 11.1 & 10.4 & 63.0 & 59.4 & \phantom{0}5.8 \\
 &  & False & \phantom{0}4.0 & \phantom{0}4.0 & \phantom{0}4.9 & \phantom{0}4.9 & 60.0 & 59.5 & \phantom{0}0.8 \\
\midrule
 \rowcolor{lightskyblue!30} \cellcolor{white}\multirow[t]{4}{*}{\texttt{Phi-3}} & \multirow[t]{2}{*}{Zero-shot} & True & \phantom{0}0.1 & \phantom{0}0.1 & \phantom{0}0.1 & \phantom{0}0.1 & 62.8 & 61.4 & \phantom{0}2.1 \\
  \rowcolor{lightskyblue!30} \cellcolor{white}& & False & \phantom{0}0.0 & \phantom{0}0.0 & \phantom{0}0.0 & \phantom{0}0.0 & 62.3 & \underline{62.3} & \phantom{0}0.1 \\
\cline{2-10}
 & \multirow[t]{2}{*}{Few-shot} & True & \underline{34.4} & 22.9 & 49.4 & 32.8 & 62.6 & 41.6 & 33.5 \\
 &  & False & 27.0 & 26.9 & 36.2 & 36.0 & 57.2 & 56.8 & \phantom{0}0.6 \\
\midrule
 \rowcolor{lightskyblue!30} \cellcolor{white}\multirow[t]{4}{*}{\texttt{Gemma-7B}} & \multirow[t]{2}{*}{Zero-shot} & True & \phantom{0}1.7 & \phantom{0}1.5 & \phantom{0}2.4 & \phantom{0}2.1 & 34.4 & 30.3 & 11.8 \\
  \rowcolor{lightskyblue!30} \cellcolor{white}&  & False & \phantom{0}0.8 & \phantom{0}0.8 & \phantom{0}2.0 & \phantom{0}2.0 & 35.3 & 35.2 & \phantom{0}0.4 \\
\cline{2-10}
 & \multirow[t]{2}{*}{Few-shot} & True & 25.0 & 17.4 & 45.3 & 31.5 & 65.8 & 45.8 & 30.5 \\
 &  & False & \phantom{0}2.8 & \phantom{0}2.5 & \phantom{0}3.3 & \phantom{0}3.0 & 32.8 & 29.9 & \phantom{0}8.9 \\
\midrule
 \rowcolor{lightskyblue!30} \cellcolor{white}\multirow[t]{4}{*}{\texttt{Gemma-2B}} & \multirow[t]{2}{*}{Zero-shot} & True & \phantom{0}1.9 & \phantom{0}1.8 & \phantom{0}4.2 & \phantom{0}4.0 & 30.6 & 29.3 & \phantom{0}4.0 \\
  \rowcolor{lightskyblue!30} \cellcolor{white}&   & False & \phantom{0}3.4 & \phantom{0}3.4 & \phantom{0}6.4 & \phantom{0}6.4 & 36.8 & 36.8 & \phantom{0}0.2 \\
\cline{2-10}
 & \multirow[t]{2}{*}{Few-shot} & True & 21.8 & 17.2 & 48.4 & 38.2 & 52.4 & 41.3 & 21.2 \\
 &  & False & 17.4 & 16.4 & 37.7 & 35.7 & 41.4 & 39.1 & \phantom{0}5.4 \\
\midrule
 \rowcolor{lightskyblue!30} \cellcolor{white}\multirow[t]{4}{*}{\texttt{RGemma-2B}} & \multirow[t]{2}{*}{Zero-shot} & True & 10.4 & 10.2 & 12.4 & 12.1 & 40.1 & 39.3 & \phantom{0}2.1 \\
  \rowcolor{lightskyblue!30} \cellcolor{white}&  & False & \phantom{0}9.9 & \phantom{0}9.9 & 10.4 & 10.4 & 40.2 & 40.0 & \phantom{0}0.5 \\
\cline{2-10}
 & \multirow[t]{2}{*}{Few-shot} & True & 33.6 & 29.4 & 45.8 & 40.0 & 50.8 & 44.5 & 12.6 \\
 &  & False & 33.6 & \underline{31.9} & 45.9 & 43.6 & 51.0 & 48.5 & \phantom{0}5.0 \\
\midrule
 \rowcolor{lightskyblue!30} \cellcolor{white}\multirow[t]{4}{*}{\texttt{Llama-3}} & \multirow[t]{2}{*}{Zero-shot} & True & 14.6 & 13.9 & 20.0 & 19.1 & 52.0 & 49.5 & \phantom{0}4.8 \\
  \rowcolor{lightskyblue!30} \cellcolor{white}&  & False & 13.4 & 13.4 & 19.3 & 19.3 & 52.1 & 52.1 & \phantom{0}0.1 \\
\cline{2-10}
 & \multirow[t]{2}{*}{Few-shot} & True & 32.1 & 31.2 & \underline{58.2} & 56.6 & 62.6 & 60.8 & \phantom{0}2.8 \\
 &  & False & 33.8 & \textbf{33.7} & 57.5 & \underline{57.3} & 60.3 & 60.0 & \phantom{0}0.4 \\
\midrule
 \rowcolor{lightskyblue!30} \cellcolor{white}\multirow[t]{4}{*}{\texttt{Llama-2}} & \multirow[t]{2}{*}{Zero-shot} & True & \phantom{0}4.4 & \phantom{0}4.2 & \phantom{0}6.5 & \phantom{0}6.2 & 57.5 & 55.0 & \phantom{0}4.5 \\
 \rowcolor{lightskyblue!30} \cellcolor{white}&   & False & \phantom{0}0.8 & \phantom{0}0.8 & \phantom{0}1.1 & \phantom{0}1.1 & 59.7 & 59.7 & \phantom{0}0.1 \\
\cline{2-10}
 & \multirow[t]{2}{*}{Few-shot} & True & 18.9 & \phantom{0}3.2 & 40.4 & \phantom{0}6.8 & 47.2 & \phantom{0}7.9 & 83.2 \\
 &  & False & 19.4 & 19.2 & 36.0 & 35.7 & 42.2 & 41.8 & \phantom{0}1.0 \\
\bottomrule
\end{tabular}
}
\caption{Precision and recall (in \%) with  exact matching (EM), alias matching (AM), and fuzzy matching (FM), for the few-shot / zero-shot models and with / without the abstain instruction. Highest scores in bold, second highest underlined.} 
\label{tab:micro_scores}
\end{table}

These observations highlight the limitations of exact matching-based evaluation, which can unfairly penalize models that tend to generate more contextual, rich, and verbose answers, like \texttt{SaulLM} and  \texttt{Mistral-7B}. Incorporating alias and fuzzy matching provides a better understanding of each model's true performance as a knowledge base.

\subsection{LLMs can be instructed to abstain from generating incorrect answers}

Table \ref{tab:micro_scores} reports the abstain rate under each prompting strategy, with or without the abstain instruction. Most models abstain more after adding the abstain instruction to the prompt. Abstaining increases precision by answering fewer questions. Interestingly, despite no few-shot examples with answer ``I don't know'', the models still abstain a significant fraction of the time.


\begin{figure}[ht]
    \includegraphics[width=0.5\textwidth]{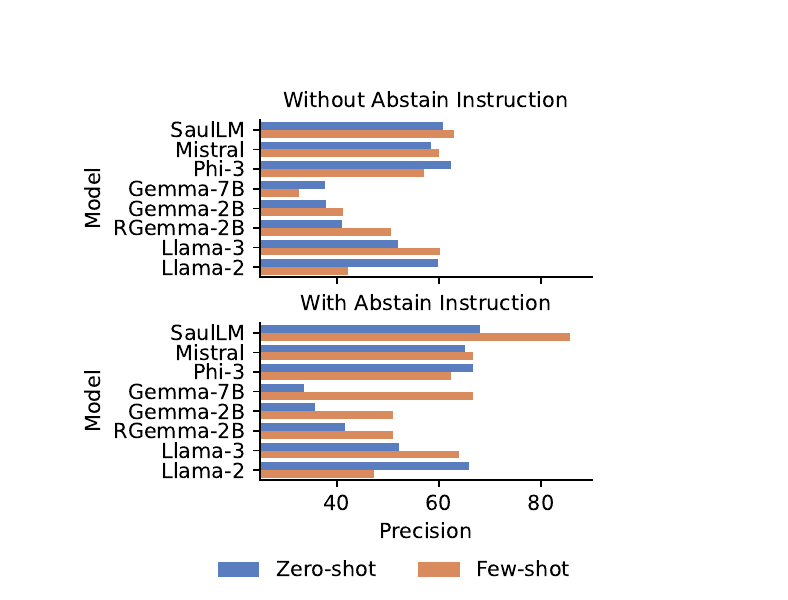}
    \caption{Comparison of precision (FM), with and without the abstain instruction, under the two prompting strategies.}
    \label{fig:precision_abstain}
\end{figure}

Including an abstain instruction generally improves precision across different models, as summarized in Figure~\ref{fig:precision_abstain}. The instruction helps the models to refrain from providing incorrect answers without needing access to the probability scores of their responses. This behavior is particularly important for LLMs accessible only through APIs that do not provide probability scores. However, recall tends to decrease. The intuitive expectation is that when models are allowed to abstain, precision goes up, and recall goes down. Our results confirm this intuition, highlighting the trade-off between precision and recall when implementing an abstain option. For sensitive applications like the legal domain, prioritizing precision over recall may be more desirable to avoid the severe consequences of incorrect information.


\subsection{In-context examples increase the factuality of LLMs}

We evaluate if in-context examples enhance the factuality of LLMs by comparing precision scores between two prompting strategies: with (few-shot) and without (zero-shot) in-context examples, as shown in Figure \ref{fig:precision_abstain}. To isolate the impact of in-context examples, we calculated precision only for questions where the model did not abstain in either setting.
Our observations indicate that in-context examples improve precision for 6 out of 8 models when instructed to abstain and for 5 out of 8 models when not prompted to abstain. The improvement is more pronounced with the abstain instruction.

These findings suggest that in-context examples can enhance the precision and, thus, the factuality of LLMs. We analyzed instances where the few-shot setting produced correct answers while the zero-shot setting did not, revealing two main benefits:

\begin{enumerate}
    \item In-context examples help the model learn the expected type and format of answers \cite{min-etal-2022-rethinking}. For instance, when asked ``applies to jurisdiction'', models in the zero-shot setting often responded with names of states. However, with in-context examples, the models correctly answered ``USA''. Similarly, for ``legislated by'', the model predicted presidents instead of the correct session of the U.S Congress for US legislation. So, the model learns a mapping between the type of legislation and the type of the answer.
    \item  In-context examples correct wrong patterns learned during pre-training. For example, case titles generally follow the format ``plaintiff v. defendant''.  Without in-context examples, all models incorrectly reversed these roles,  which was corrected with few-shot prompts. This ability to override wrong prior patterns is explained in \cite{wei2023larger}.
\end{enumerate}

\noindent To summarize, in-context examples significantly enhance the precision and factuality of LLMs by guiding them on the expected type and format of answers and correcting erroneous patterns from pre-training.

\subsection{Training an LLM on legal documents improves its factuality}

Finally, we investigate whether training an LLM on legal documents improves its factuality by comparing the performance of \texttt{SaulLM}, which is \texttt{Mistral-7B} with additional pre-training on legal documents, against the standard \texttt{Mistral-7B} model. The top performance was achieved by \texttt{SaulLM}, outperforming \texttt{Mistral-7B}. Under the few-shot prompt with the instruction to abstain, \texttt{SaulLM} achieved a precision of 81\%, significantly higher than the 63\% precision of \texttt{Mistral-7B}. This suggests that training \texttt{SaulLM} on legal documents enhances its ability to provide correct answers, similar to the effect of few-shot prompting. The additional training helps \texttt{SaulLM} understand the context and format of legal questions more accurately, leading to better performance.
Thus, training an LLM on legal documents substantially improves its factuality, particularly in terms of precision. This improvement aligns with the benefits observed from few-shot prompting, indicating that domain-specific training enables the model to generate more accurate answers.

Interestingly, while \texttt{SaulLM} showed a significant increase in precision, its recall did not improve as markedly. This can be explained by the much higher abstain rate of \texttt{SaulLM} compared to Mistral-7B.


\section{Conclusions}
\label{sec:conclusion}

In this paper, we explored the factuality of LLMs as KBs in the legal domain. We evaluated various models, including \texttt{SaulLM}, pre-trained on legal documents.

Our findings reveal that the performance of LLMs improves significantly when using alias and fuzzy matching instead of exact matching. Abstain instructions and few shot prompting increase factuality. Pre-training on legal documents, as shown by \texttt{SaulLM}, substantially improves the precision, highlighting the importance of domain-specific pre-training.

Our careful evaluation methods reveal that LLMs hallucinate significantly less than reported in \cite{dahl_large_2024}, due to our consideration of the verbose nature of LLM responses. While \texttt{SaulLM} achieves a high precision of 81\%, this is still insufficient for high-stakes legal applications, but may suffice for research and analytics. Instances of lawyers being sanctioned for using fictitious cases underscore the risks of factual errors.

These results highlight the potential of LLMs as supplementary tools for legal research, emphasizing the importance of accurate atomic facts. Future work should refine evaluation methods and explore domain-specific training to enhance LLM reliability.

\section*{Acknowledgements}
This research was supported by funding from Labex DigiCosme, which is gratefully acknowledged.

\bibliographystyle{ACM-Reference-Format}
\balance
\bibliography{references, manual_references}

\end{document}